%% file: acl_latex.tex
\pdfoutput=1

\documentclass[11pt]{article}

\usepackage[final]{acl}

\usepackage{times}
\usepackage{latexsym}
\usepackage{microtype}
\usepackage{hyperref}
\usepackage{url}
\usepackage{booktabs}
\usepackage{graphicx}

\usepackage{subfigure}
\usepackage{enumitem}
\newenvironment{itemize*}%
 {\leftmargini=20pt\begin{itemize}%
  \setlength{\itemsep}{3pt}%
  \setlength{\parskip}{0pt}%
  }%
 {\end{itemize}}
\newenvironment{enumerate*}%
 {\begin{enumerate}%
  \setlength{\itemsep}{0pt}%
  \setlength{\parskip}{0pt}}%
 {\end{enumerate}}

\usepackage{cleveref}
\usepackage{listings}
\usepackage{titlesec}
\usepackage{multirow}
\usepackage{makecell}
\usepackage{xcolor}

\usepackage{perpage}
\MakePerPage{footnote}

\definecolor{lightred}{RGB}{255,163,163}
\definecolor{deepred}{RGB}{146,0,0}
\newcommand{\lightred}{\textcolor{lightred}}
\newcommand{\deepred}{\textcolor{deepred}}

\usepackage[T1]{fontenc}

\usepackage[utf8]{inputenc}

\usepackage{microtype}

\usepackage{inconsolata}

\title{Tell Me More! Towards Implicit User Intention Understanding of Language Model Driven Agents}

\author{Cheng Qian$^{1}$\thanks{indicates equal contribution.},  Bingxiang He$^{1*}$,  Zhong Zhuang$^{1}$,  Jia Deng$^{2}$,  Yujia Qin$^{1}$, Xin Cong$^{1}$,  \\
\textbf{Zhong Zhang$^{1}$, Jie Zhou$^{3}$, Yankai Lin$^{2}$, Zhiyuan Liu$^{1}$, Maosong Sun$^{1}$}\\
$^{1}$Tsinghua University,  \hspace{0.3em}$^{2}$Renmin University of China,  \hspace{0.3em}$^{3}$WeChat AI, Tencent Inc.\\
\texttt{\{qianc20, hbx20\}@mails.tsinghua.edu.cn}\\}

\begin{document}

\maketitle
\input{Sections/0_Abstract}
\input{Sections/1_Introduction}
\input{Sections/2_Related_Works}
\input{Sections/3_Dataset}
\input{Sections/3_Method}
\input{Sections/4_Experiments}
\input{Sections/5_Discussions}
\input{Sections/6_Conclusion}

\bibliography{anthology,custom}

\input{Sections/Appendix}

\end{document}

%% file: Sections/0_Abstract.tex
\begin{abstract}
Current language model-driven agents often lack mechanisms for effective user participation, which is crucial given the vagueness commonly found in user instructions. Although adept at devising strategies and performing tasks, these agents struggle with seeking clarification and grasping precise user intentions.
To bridge this gap, we introduce Intention-in-Interaction (IN3), a novel benchmark designed to inspect users' implicit intentions through explicit queries. Next, we propose the incorporation of model experts as the upstream in agent designs to enhance user-agent interaction. Employing IN3, we empirically train Mistral-Interact, a powerful model that proactively assesses task vagueness, inquires user intentions, and refines them into actionable goals before starting downstream agent task execution. Integrating it into the XAgent framework, we comprehensively evaluate the enhanced agent system regarding user instruction understanding and execution, revealing that our approach notably excels at identifying vague user tasks, recovering and summarizing critical missing information, setting precise and necessary agent execution goals, and minimizing redundant tool usage, thus boosting overall efficiency.
All the data and codes are released\footnote{https://github.com/HBX-hbx/Mistral-Interact}.
\end{abstract}

%% file: Sections/1_Introduction.tex
\section{Introduction}

Large language models including the OpenAI GPT \citep{openai2022chatgpt, openai2023gpt4}, LLaMA \citep{touvron2023llama, touvron2023llama2}, and Mistral series \citep{jiang2023mistral} have made great strides in high-quality text and code generation \citep{zeng2022glm, chowdhery2022palm, openai2023gpt4, touvron2023llama2}, complex logical reasonings \citep{wei2023chainofthought, gao2023pal, yao2022react, yao2023tree}, and using external tools \citep{schick2023toolformer, qin2023tool, qin2023toolllm}. These traits enable the language model to interact with the outside world and receive feedback as an AI agent to assist user tasks \citep{xi2023rise, wang2023survey}, inspiring many open-source frameworks focusing on agent design, including BabyAGI \citep{babyagi2023} AutoGen \citep{wu2023autogen}, CAMEL \citep{li2023camel}, AutoGPT \citep{autogpt2023}, and XAgent \citep{xagent2023}, etc. These frameworks generally leverage the backbone language model's exceptional abilities to comprehend user instructions and execute user tasks.

\begin{figure}[!t]
    \centering
    \subfigure{\includegraphics[width=\linewidth]{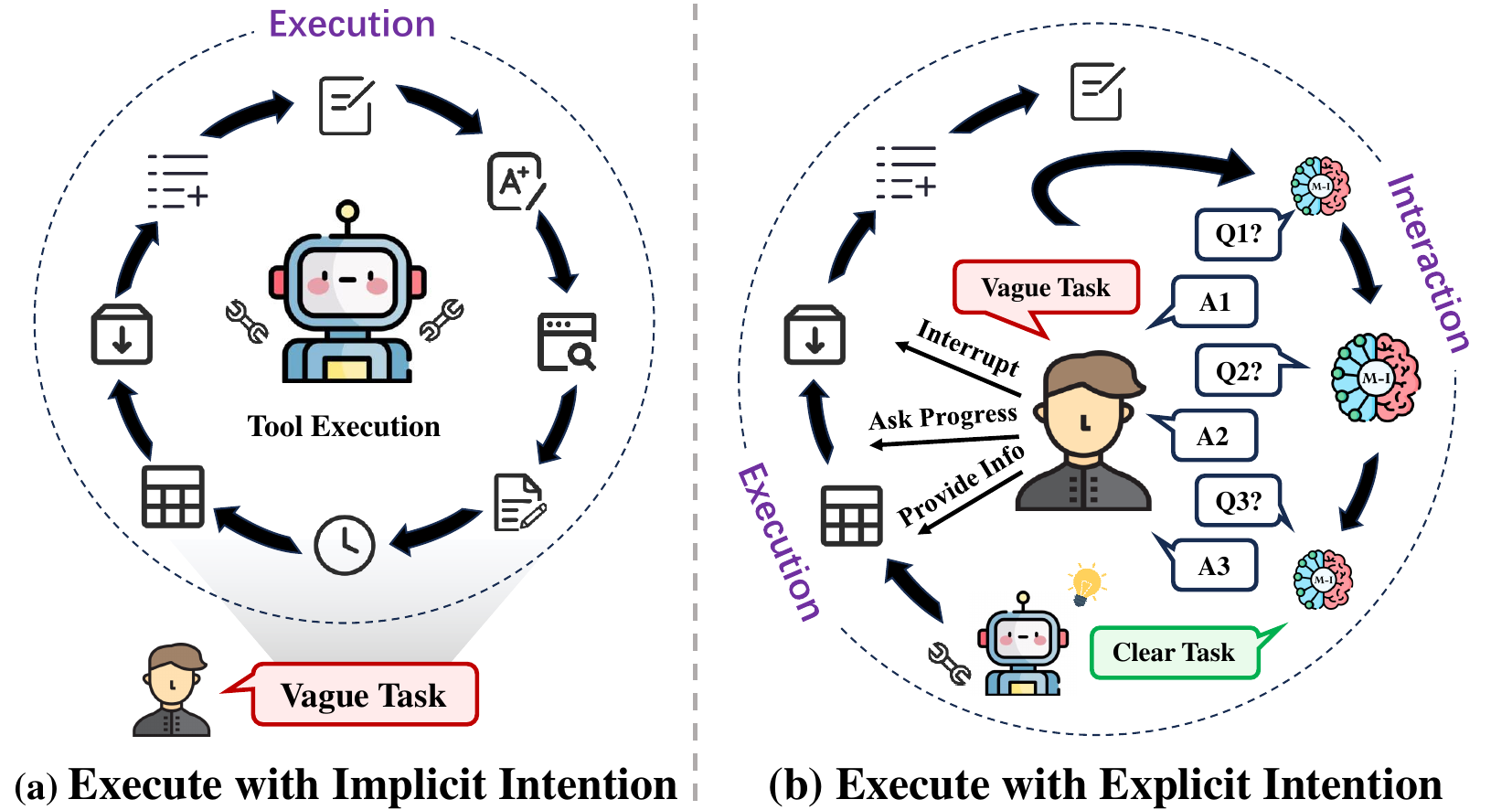}}
    \caption{A comparison of agent execution with implicit intentions or explicit intentions after user-agent interaction.}
    \label{fig:intro}
\end{figure}

However, current agent designs seldom consider robust \textbf{user interaction}, despite that i) the user's initial instruction to the agent system is usually vague and brief, and ii) different users have different intentions which necessitate explicit query and inspiration. The ignorance of the user's clear and specific needs often leads to ``fake success'' during agent task execution, where it seemingly completes the goal but deviates far from the user's true intentions. This stresses the importance of \textbf{implicit intention understanding} during user-agent interaction to further improve the agent design's robustness and efficiency.

Current agent benchmarks usually assume the clearance of given tasks and exclude user intention understanding as an important aspect for evaluation. Given this ignorance in assessment, we formulate \textbf{Intention-in-Interaction} (\textbf{IN3}), a benchmark aiming to test the agent's interaction ability through explicit task vagueness judgment and user intention understanding. IN3 provides general agent tasks over hundreds of categories, each with its vagueness, missing details, and each detail's importance level and options to inspire the user's true intentions.
In addition, we propose an innovative approach to evaluate present agent systems, with metrics addressing the existing gap in evaluation by incorporating two key aspects that emphasize user participation:
\begin{itemize}[topsep=0pt, partopsep=1pt, leftmargin=12pt, itemsep=0pt]
\item \textbf{Instruction Understanding}: To \emph{directly} evaluate the process of intention understanding, we quantitatively measure the preciseness of the agent in judging instruction's vagueness, the recovery rate of important missing details, and the robustness of user intention summarization under diverse scenarios.
\item \textbf{Instruction Execution}: To \emph{indirectly} reflect the effectiveness of intention understanding, we contrast the agent task execution given the initial (vague) or finalized (clear) user goal, and measure the generality, necessity, and efficiency of the agent's actions.
\end{itemize}

As language models lie as the core of agent designs, we first apply sampled tasks from IN3 to test the active interaction ability of diverse models, revealing that most of them seriously lack the ability to judge task vagueness and initiative to comprehensively understand user intentions. This further raises our research problem: \textbf{How to enhance the interaction ability of the agent system through the underlying model's robust intention understanding?}

To address this, we propose to integrate a specialized upstream model in charge of user-agent interaction in agent design. As illustrated in \Cref{fig:intro}, to explicitly understand user's specific intentions, the model should actively and explicitly ask users for missing details before passing the task for downstream execution. To enhance this interaction capability, we apply training split tasks in IN3 to construct simulated model-user conversation records that provide explicit initial thoughts, rounds of queries with options, summarization of implicit intentions, and diverse user response tones. Training on these conversations, we adapt Mistral-7B into \textbf{Mistral-Interact}, a powerful and robust variant capable of judging the vagueness of user instruction, actively querying for missing details with suggestions, and explicitly summarizing the detailed and clear user intentions.

In experiments, we incorporate Mistral-Interact into the XAgent framework \citep{xagent2023}, an autonomous agent system for complex task solving. Through comprehensive evaluation of both user instruction understanding and agent instruction execution, we reveal that our adapted Mistral-Interact can i) correctly judge the vagueness of over 85\% tasks, recover over 70\% of the most important missing details, and summarize over 96\% implicit user intentions without omission, ii) significantly reduce the number of too general or unnecessary goals and lower the tool invocation times during agent execution, which raises overall agent efficiency.
The performance of Mistral-Interact is more aligned with human preferences, far better than the previous LLaMA and Mistral series while rivaling closed-source GPT-4 but with a much smaller scale. Our method proves the viability of integrating smaller, open-sourced model experts for robust implicit intention understanding during user-agent interaction, and we conclude by discussing its further implications and future directions.
Overall, our key contributions include:
\begin{itemize}[topsep=0pt, partopsep=1pt, leftmargin=12pt, itemsep=0pt]
\item We formulate a new research question regarding the enhancement of user-agent interaction through robust intention understanding, and release the IN3 benchmark that focuses on user participation within current agent designs.
\item We propose to integrate an expert specialized in interaction as an upstream design before task execution in the agent system, and empirically adapt Mistral-Interact, shifting the focus from complex agent design to smaller-scale user-centric module design.
\item We create a set of new metrics regarding the evaluation of user-agent interaction, which takes into consideration both the quantifiability of results and alignment to user preferences for future benchmarks to follow.
\item We prove the viability of our method through comprehensive experiments and case studies on the XAgent framework, thereby promoting a new mechanism and paradigm of user-agent interaction in agent designs.
\end{itemize}

%% file: Sections/2_Related_Works.tex
\section{Related Works}

\paragraph{LLM-driven Agent.} Recent large language models (LLMs), including the closed-source GPT series \citep{openai2022chatgpt, openai2023gpt4} and open-source LLaMA \citep{touvron2023llama, touvron2023llama2}, Mistral \citep{jiang2023mistral} series, have demonstrated strong reasoning \citep{wei2023chainofthought, gao2023pal, yao2022react, shinn2023reflexion}, planning \citep{yao2023tree, besta2023graph, sel2023algorithm, hao2023reasoning, ye2023large} and tool using ability \citep{nakano2021webgpt, huang2022language, ahn2022can, schick2023toolformer, patil2023gorilla, qin2023tool, qin2023toolllm, qian2023creator, qian2023toolink}. These enable LLMs to interact with the world as AI agents, accomplishing complex and grounded human tasks \citep{xi2023rise, wang2023survey, babyagi2023, autogpt2023, li2023camel, wu2023autogen, xagent2023}. A line of current research focused on enhancing agent's ability through tool creation \citep{cai2023large, qian2023creator, wang2023voyager}, multimodal capability \citep{gupta2023visual, shen2023hugginggpt, wu2023visual}, and domain-specific tools and resources \citep{jin2023genegpt, lyu2023gitagent, ye2023proagent}. Other studies involve multi-agent frameworks for communication \citep{park2023generative, li2023camel, qian2023communicative, hu2023chatdb}, collaboration \citep{chen2023agentverse, wu2023autogen}, and evaluation \citep{chan2023chateval, zhang2023wider}. These previous works have not fully taken into account the role of \emph{users} in the agent designs, the issue that our study strives to address.

\paragraph{User Intention Understanding.} In the quest to develop an engaging agent, it becomes paramount to anticipate user intentions across varied contexts. Conventional approaches for user intention understanding include Support Vector Machine (SVM) \citep{Sullivan2018Comparing}, Naive Bayes \citep{Vikramkumar2014Bayes} and XGBoost \citep{chen2016XGBoost}, with XGBoost consistently showing superior performance \citep{Cai2020Predicting}. Accurate user intention understanding is crucial, especially in information-seeking scenarios like web search engines or community question-answering platforms. The MSDialog dataset \citep{qu2018analyzing} is specifically crafted to dissect user intention distribution, co-occurrence, and flow patterns, while the SHDCRF model \citep{shen2011sparse} is adept at learning from user search sessions. Moreover, categorizing user utterances based on hand-crafted features, such as leveraging Wikipedia \citep{hu2009understanding}, textual and metadata features \citep{chen2012understanding}, and factors like content, discourse, sentiment, and context \citep{Cai2020Predicting}, can significantly augment user intention understanding.
With the rise of language models, now the understanding of user intention can be achieved through zero-shot prompting \citep{kotnis2022humancentric}, enabling the recommendation of task-oriented bots based on user intentions \citep{ kuo2023ZeroShotPrompting}. In contrast, our work is the first to center on incorporating implicit user intention understanding within \emph{agent designs}.

\paragraph{User-Agent Interaction Designs.} Current agents often struggle with intricate reasoning tied to social and historical contexts, and they face challenges in adapting to situations requiring personalization \citep{Majumder2023UCNLP}. A user-centric approach to agent design emphasizes alignment with user preferences and fosters effective interactions. One research direction advocates for agents as teammates, involving stakeholders, developers, and designers throughout agent designs to incorporate authentic human requirements \citep{kotnis2022humancentric}. Another avenue explores the role of agents in fostering creativity, leveraging their creative perspective to enhance communication and interaction with users in a manner resembling human engagement \citep{Weiwen2020InteractiveAI, Zhuohao2021Human-AI}. An extension of this approach includes treating agents as collaborative team members, considering and assigning personality traits and skills necessary for an agent to be recognized and accepted as part of the team \citep{Christina2021TeammateAgent}. Different from these works, we focus on the user's \emph{intention understanding} during interaction to enhance the agent's effectiveness and efficiency.

%% file: Sections/3_Dataset.tex
\section{Intention-in-Interaction Benchmark}
\label{sec:in3_dataset}
Most of the past agent benchmarks assume the given task is clear, aiming to evaluate the agent's execution ability. However, user-provided instructions are often ambiguous. For instance, for the task ``Locate the best yoga class in my city'' presented in \Cref{fig:dataset}, the instruction is unclear about where ``my city'' is and what the criteria for the ``best'' is. 
All these vagueness necessitate a clearer comprehension of the user's true intentions to further boost the agent's execution efficiency.

To this end, we aim to formulate a more rational agent task setting, in which the user's true intentions for each task are implicit. The completion of these tasks requires the agent to actively inquire about missing details and understand the user's implicit intentions.
To comprehensively enhance and quantitatively evaluate these abilities in agent designs, we introduce \textbf{Intention-in-Interaction} \textbf{(IN3)}, a benchmark striving to assess and inspire the agent's intention understanding ability in a robust, structured, and user-friendly way.

\subsection{Benchmark Construction}

\begin{figure}[!t]
    \centering
    \subfigure{\includegraphics[width=\linewidth]{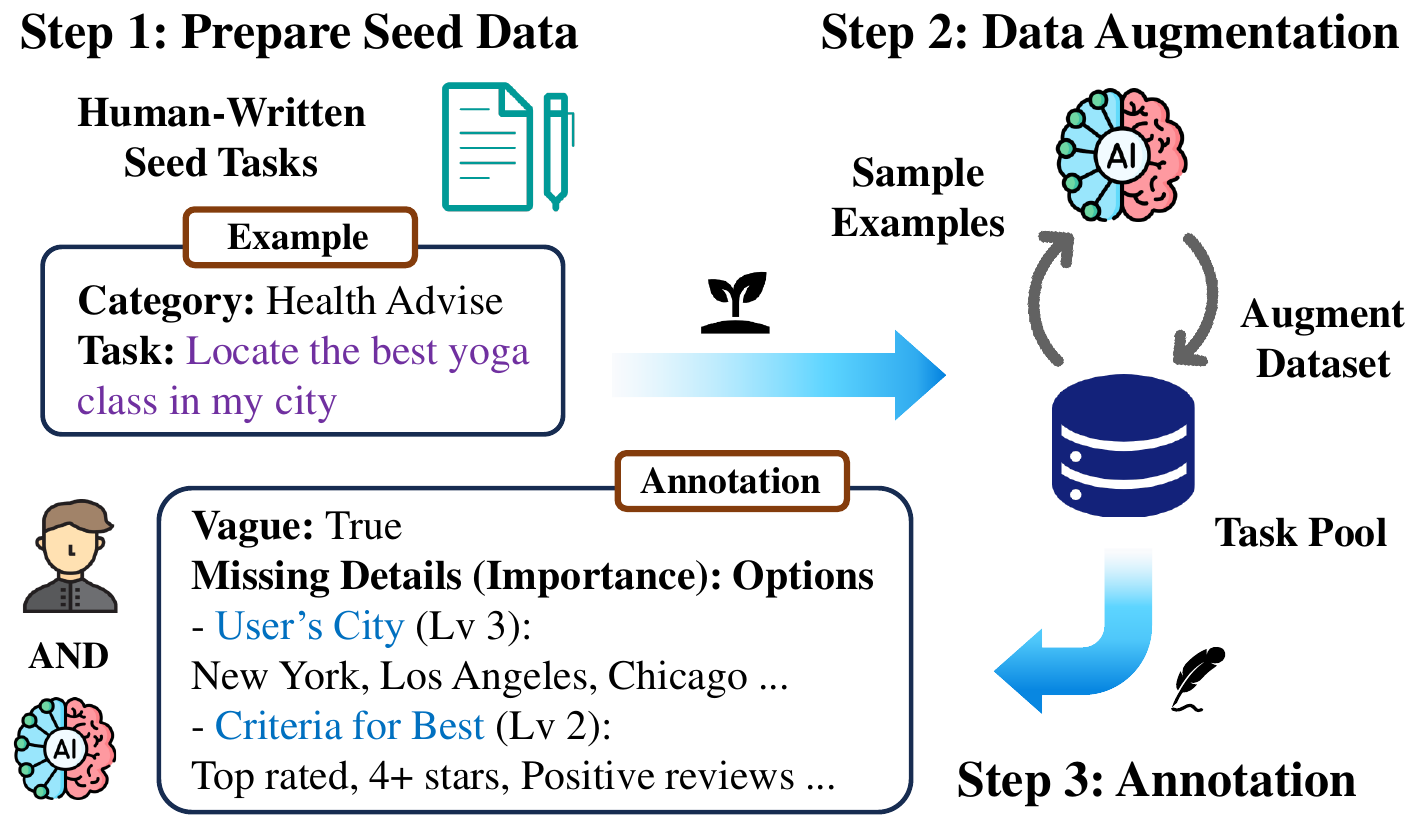}}
    \caption{An illustration of IN3's formation with an example data point.}
    \label{fig:dataset}
\end{figure}

\begin{table}[!t]
\centering
\footnotesize
\begin{tabular}{lcc}
\toprule
\textbf{\quad\quad Split} & \textbf{Training} & \textbf{Test}\\
\midrule
\begin{tabular}[c]{@{}l@{}}\textbf{Task}\\ \quad\quad- \textit{Vague}\\ \quad\quad- \textit{Clear} \end{tabular} 
& \begin{tabular}[c]{@{}c@{}}1261\\ 1012\\ 249 \end{tabular}
& \begin{tabular}[c]{@{}c@{}}108\\ 95\\ 13 \end{tabular} \\
\midrule
\textbf{Categories} & 250 & 50 \\
\midrule
\begin{tabular}[c]{@{}l@{}}\textbf{\# Missing Details}\\ \quad\quad- \textit{Avg.}\\ \quad\quad- \textit{Lv 1} (\%)\\ \quad\quad- \textit{Lv 2} (\%)\\ \quad\quad- \textit{Lv 3} (\%)\end{tabular}
& \begin{tabular}[c]{@{}c@{}}3615\\ 3.57\\ 15.44\\ 67.75\\ 16.81\end{tabular}
& \begin{tabular}[c]{@{}c@{}}350\\ 3.68\\ 9.14\\ 72.29\\ 18.57\end{tabular} \\
\midrule
\begin{tabular}[c]{@{}l@{}}\textbf{\# Options}\\ \quad\quad- \textit{Avg.}\end{tabular}
& \begin{tabular}[c]{@{}c@{}}11523\\ 11.39\end{tabular}
& \begin{tabular}[c]{@{}c@{}}1042\\ 10.97\end{tabular} \\
\bottomrule
\end{tabular}
\caption{Detailed statistics for training and test splits of IN3. The missing details and option numbers are averaged on the number of vague tasks.}
\label{tab:dataset_statistics}
\end{table}

IN3 provides diverse agent tasks over hundreds of categories (e.g. Cookery, Arts, Programming), with annotations on whether the task is vague, what the missing details are (if vague), the importance level of each missing detail (three levels, the higher the more important, detailed in \Cref{apdx:in3_annotation}), and the potential options for each missing detail. Regarding the vague health advice task presented in \Cref{fig:dataset}, IN3 provides annotated missing details about the user's city and criteria for the best, with options to demonstrate potential answers and inspire the user's true intention. Since the city is indispensable for yoga class searching, it has a higher importance level (Lv 3) than the missing criteria (Lv 2), which only serves to better match the user's preference.

The task description and its category are generated in a self-instruct manner \citep{wang2023selfinstruct} applying GPT-4. We in total consider over 200 categories and construct 1300+ diverse agent tasks. As illustrated in \Cref{fig:dataset}, with human-written seed tasks (Step 1), the model iteratively generates new tasks to augment the dataset, while sampling demonstrations from the dataset as new examples for itself to perform the next round of generation (Step 2), detailed in \Cref{apdx:in3_task}. We perform human annotation of each task's vagueness, missing details, and each detail's importance level and potential options with the help of GPT-4 (Step 3). GPT-4 will first suggest the task's vagueness and potential missing details with options and importance level, while human annotators take them as references and adapt them with their own perspectives and intentions, detailed in~\Cref{apdx:in3_annotation}.

Overall, IN3 can be applied to evaluate the agent's discernment of task vagueness, assess the agent's ability to recover important missing details, and facilitate training on the underlying model's implicit intention understanding capability. IN3 is also divided into training and test splits, with more statistical details provided in~\Cref{tab:dataset_statistics}.

\begin{table*}[!t]
\begin{center}
\footnotesize
\tabcolsep=0.025\linewidth
\begin{tabular}{clccc}
\toprule
\textbf{(Quantitative)} & \textbf{Metric} & \textbf{LLaMA-2}  & \textbf{Mistral} & \textbf{GPT-4}  \\
\midrule
\textbf{\makecell{Vagueness\\Judgment}}   & \textit{Judgment accuracy}                  & 6 / 10        & 8 / 10       & 9 / 10  \\
\midrule
\textbf{\makecell{Missing\\Details}}       & \textit{Ratio of necessary query attempts} & 7 / 23        & 7 / 8        & 16 / 24 \\
\toprule
\textbf{(Qualitative)} & \textbf{Description} & \textbf{LLaMA-2}  & \textbf{Mistral} & \textbf{GPT-4}  \\
\midrule
\multirow{2}{*}{\textbf{\makecell{During\\Judging}}}
& \textit{Inaccuracy in judgment}                          & \textbf{\deepred{Severe}}        & \textbf{\lightred{Light}}        &         \\
& \textit{Judge as clear but still query}     & \textbf{\lightred{Light}}         &              &         \\
\midrule
\multirow{5}{*}{\textbf{\makecell{During\\Inquiring}}}
& \textit{Ask too much without stopping}                          & \textbf{\deepred{Severe}} &          &         \\
& \textit{Ask too few, quickly jump to summarize}                          &     
          & \textbf{\deepred{Severe}}&         \\
& \textit{Inquire unnecessary details}                      & \textbf{\deepred{Severe}}        &              & \textbf{\lightred{Light}}   \\
& \textit{Too many queries at one time}                     & \textbf{\lightred{Light}}         & \textbf{\lightred{Light}}        & \textbf{\lightred{Light}}   \\
& \textit{Query all over one aspect, ignoring others}   & \textbf{\lightred{Light}}         &              & \textbf{\lightred{Light}}   \\
\midrule
\multirow{3}{*}{\textbf{\makecell{During\\Summarizing}}}
& \textit{Solve task instead of provide summary}            & \textbf{\deepred{Severe}}       &              &         \\
& \textit{Provide assumed or hallucinated details}          & \textbf{\lightred{Light}}         & \textbf{\lightred{Light}}        & \textbf{\lightred{Light}}   \\
& \textit{Ignore user provided details}                       & \textbf{\lightred{Light}}         &              &        \\
\bottomrule
\end{tabular}
\caption{The performance and challenges of LLaMA-2-7B-Chat, Mistral-7B-Instruct-v0.2, and GPT-4's intention understanding ability during interaction.}
\label{tab:preliminary_evaluation}
\end{center}
\end{table*}

\subsection{Preliminary Test}
\label{sec:preliminary_study}
\paragraph{Settings.} As the language model lies at the core of agent designs, we begin by conducting a preliminary study on current open-source and closed-source models' intention understanding ability during interaction. Specifically, we sample ten tasks from IN3 and apply them to test LLaMA-2-7B-Chat\footnote{https://huggingface.co/meta-llama/Llama-2-7b-chat-hf}, Mistral-7B-Instruct-v0.2\footnote{https://huggingface.co/mistralai/Mistral-7B-Instruct-v0.2}, and GPT-4. All models are prompted with the same instructions to i) judge the vagueness of the task, ii) inquire the user for missing details if the task is vague, and iii) summarize the detailed user task goal. Please refer to \Cref{apdx:preliminary_exp} for specific tasks, detailed settings, and prompt contents.

\paragraph{Evaluation.} For \textbf{quantitative} analysis, we first count how many model judgments of task vagueness are aligned with IN3's human annotation. Next, we count the total query attempts the model made in the vague tasks, and how many of them are aligned with IN3's human annotation (which are truly meaningful, necessary, and important missing details; in total there are 24 such human-annotated missing details, shown in \Cref{apdx:preliminary_exp}). For \textbf{qualitative} analysis, we divide the whole interaction into three phases: vagueness judging, missing details inquiring, and detailed user goal summarizing. For each stage, we gather various unsatisfactory or failed interaction patterns and evaluate whether they are severe, light, or nearly nonexistent, with criteria detailed in \Cref{apdx:preliminary_eval}.

\paragraph{Results.} We present a results overview in \Cref{tab:preliminary_evaluation}. All three models present challenges in robust user intention understanding but with different failure patterns. Among them, LLaMA-2 performs the worst as it cannot identify already clear goals and often asks for unnecessary details without stopping. Mistral is a little better but still suffers from an insufficient understanding of human intentions. GPT-4 aligns the closest with human intentions regarding task vagueness and important missing details. We further provide case studies in \Cref{apdx:preliminary_case} for illustration. Our results bear broad implications:
\begin{itemize}[topsep=0pt, partopsep=1pt, leftmargin=12pt, itemsep=0pt]
\item \textbf{Shortages to Focus}: From current failed patterns, future language model-driven agent design should focus on i) making more precise judgments of task vagueness, ii) querying important missing details of diverse aspects in a user-friendly way, and iii) summarizing user preferences comprehensively without omissions. 
\item \textbf{Necessity of Adaptation}: To further raise the agent's intention understanding ability during interactions, we reveal that prompt engineering is insufficient, which necessitates further training to build interaction experts. According to the results, Mistral has generally fewer defects during the interaction, making it a better base model choice for further adaptation in agent system design.
\item \textbf{GPT-4's Ability}: GPT-4's vagueness judgment aligns most closely with users, and it could recover most of the necessary missing details regarded by users. These justify the use of GPT-4 to imitate users with specific preferences to construct conversation records for training.
\end{itemize}

\subsection{Research Problem}
According to the challenges we identify in implicit intention understanding, we formulate our research problem as follows:
For each task $t$ with a set of missing details $D = $ {\small$\{d_1, d_2, \ldots, d_n\}$}, we would like the agent to transform $t$ into $t_{user}$ for execution, where $t_{user}$ encompasses all the user's implicit intentions towards each missing detail.
This necessitates \textbf{enhancement of the agent's interaction ability through the underlying model’s robust implicit intention understanding}.

%% file: Sections/3_Method.tex
\section{Method}
To further enhance the implicit intention understanding ability of current agent designs, we propose to train a model expert specialized in implicit intention understanding through conversation with users and incorporate it as an upstream module in agent design. Denote agent execution process as a function $f$, then this module should act as a ``buffer'' between initial user task $t$ and downstream execution $f(t)$.
If $t$ is already clear, the module should directly pass $t$ for execution, while if $t$ is vague, the module should robustly chat with the user, turning $t$ into $t_{user}$ with specific user intentions.
We expect $f(t_{user})$ to perform better and be more aligned with the user's intentions than $f(t)$.

To realize this, we first apply IN3 to construct conversation records for training. Using the constructed interaction data, we adapt Mistral-7B into Mistral-Interact, a powerful variant capable of judging the vagueness of user instruction, actively querying for missing details with suggestions, and explicitly summarizing detailed user intentions. In this section, we illustrate our method in detail.

\begin{figure*}[!t]
    \centering
    \subfigure{\includegraphics[width=0.8\linewidth]{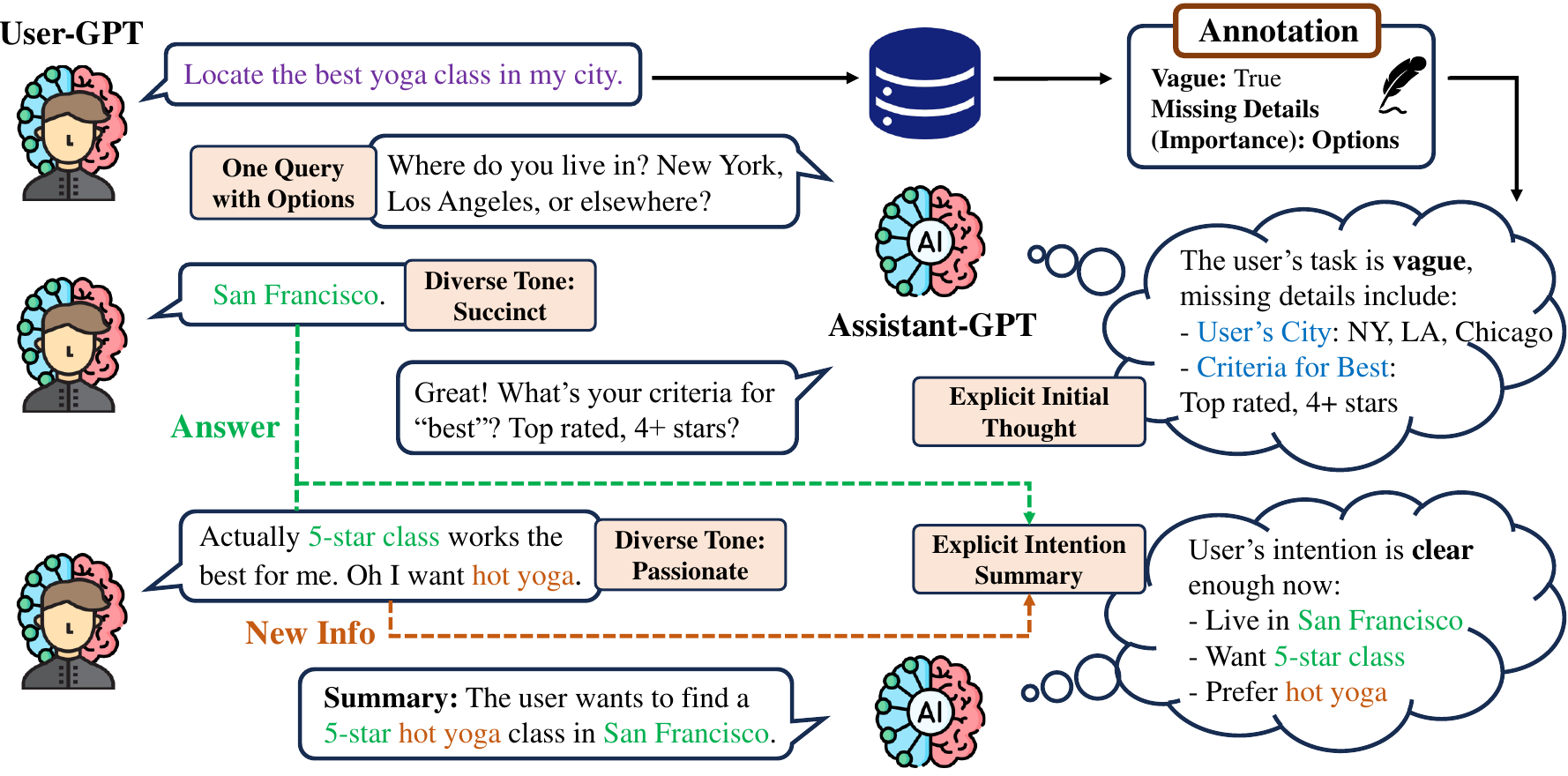}}
    \caption{The construction of conversation records with diverse strategies applied.}
    \label{fig:method}
\end{figure*}

\subsection{Construction of Training Data}
To enhance the model's implicit intention understanding through interaction, we need to further train it on how to inquire about the missing details in a vague user task through conversation.
As IN3 has already provided diverse agent tasks with annotations, we apply its training split to construct the conversation records for training. Results from the preliminary study have justified the use of GPT-4 to imitate user preferences. Therefore, to make the construction process automatic and efficient, we employ two GPT-4s to simulate the conversation, with one imitating the user aiming to complete a certain task (User-GPT), and the other as an assistant aiming to clearly understand user intentions with the annotations from IN3 as help (Assistant-GPT).

\paragraph{Strategies.} With IN3's annotations regarding task vagueness, missing details, and potential options, we apply several strategies during the construction of conversation records to better inspire the target model's robust inquiry and reasoning ability. All strategies are illustrated with an example in \Cref{fig:method}.
\begin{itemize}[topsep=0pt, partopsep=1pt, leftmargin=12pt, itemsep=0pt]
\item \textbf{Explicit Initial Thought}: After the User-GPT presents a task $t$ in IN3, we first manually construct the Assistant-GPT's initial thought with an explicit judgment of task vagueness, missing details, and potential options as presented in IN3. We concatenate all the information through a template detailed in \Cref{apdx:dataset_conversation}. This serves to guide the model to later inquire about core details instead of asking arbitrarily with reasoning on the fly.
\item \textbf{Query with Options}: For each round of conversation, we instruct the Assistant-GPT to provide thoughts and ask only \emph{one} query with options. The inquiry is made based on the missing details and options listed in the initial thought. One query at a time makes model inquiries less pushy, while reasonable options provided can inspire the user's deeper thoughts and encourage explicit expression of implicit intentions. All of these serve to make the model's interaction more user-friendly.
\item \textbf{Diverse User Tones}: For the user response in each round of conversation, we prompt User-GPT to imitate users with different tones, focusing mainly on succinct (assume users are lazy and provide short responses) and passionate (assume users provide long responses with new information) response patterns. This serves to facilitate the applicability and robustness of the models trained on it.
\item \textbf{Explicit Intention Summary}: We instruct Assistant-GPT to explicitly summarize all user-provided intentions (both responses to inquiries and new information) in thoughts and provide a clearer version of user goal $t_{user}$ once it believes enough information is gathered. The thought serves to make the summarized $t_{user}$ more logical and comprehensive without omission. Moreover, the summarized $t_{user}$ with user intentions could be directly applied for downstream agent execution $f(t_{user})$, thus promoting a seamless integration of the trained model into existing agent frameworks.
\end{itemize}

Note that if the task $t$ in IN3 is clear, then all the conversation simulations will be omitted. Only the judgment of vagueness (clear) in the initial thought and the summary thought will be performed, with the finalized $t_{user} = t$. In addition, as we imitate different user tones, the total conversation records contribute to approximately twice the training task number in IN3. Please refer to \Cref{apdx:dataset_conversation} for more details.

\subsection{Training Details}
Generally, we apply the conversation records constructed to train Mistral-7B\footnote{https://huggingface.co/mistralai/Mistral-7B-v0.1} into Mistral-Interact, aiming to make the model better comprehend user intentions and transform vague tasks into clear initiatives for the agent's execution. Specifically, we fit each conversation record into a template outlined in \Cref{Apdx:training_details} and cumulatively concatenate multiple rounds of interactions, thus creating multiple data instances.

For \emph{vague} tasks, the first data instance trains the model to generate an initial thought, judge the task's vagueness, and perform an initial inquiry if the task is vague. Then, the $i$-th data instance trains the model to generate the $i$-th inquiry on missing details with thought. Finally, the last data instance trains the model to generate a summary of the user's responses with thought.
For \emph{clear} tasks, the model is directly trained to generate thoughts, judgment, and the final summary. This involves only one data instance.

We utilize the model-center framework \citep{modelcenter2023} to conduct full-parameter fine-tuning of Mistral-7B on two 80GB A800s. Specific hyper-parameters are detailed in \Cref{apdx:method_hyperpara}.

%% file: Sections/4_Experiments.tex
\section{Experiments}
 
An agent's intention understanding capability can be assessed \emph{directly} through user interaction and \emph{indirectly} through downstream task execution. Interaction focuses on intention understanding itself, while execution focuses on intention understanding's ultimate goal, which is to enhance the agent design's efficiency.

Therefore, to comprehensively evaluate the effectiveness of an agent design capable of interaction, we divide our experiments into two aspects: i) \textbf{Instruction Understanding}: The evaluation of agent's intention understanding capability during user-agent interaction to directly show its excellence; ii) \textbf{Instruction Execution}: The evaluation of agent task performance with an upstream plug-in interaction expert to reflect its effectiveness.

\subsection{Evaluation on Instruction Understanding}
Instruction understanding does not involve any real-time agent execution, so we directly evaluate the language models themselves during interaction to judge their capability to serve as a robust upstream module in agent design.

\subsubsection{Experimental Settings}
\paragraph{Data and Settings.} We use the test split of IN3 agent tasks for evaluation. For each task, we engage the user in an open-ended conversation with the target model, during which it will actively inquire about user intentions. We recruit diverse undergraduate-level users of different expertise to participate in the conversation and provide responses, detailed in \Cref{apdx:interact_exp_user_participation}. The whole conversation process is recorded and then evaluated based on the ground truth provided in IN3.

\paragraph{Model and Baselines.} We compare Mistral-Interact with LLaMA-2-7B-Chat, Mistral-7B-Instruct-v0.2, and GPT-4, the same models in \Cref{sec:preliminary_study}. For a fair comparison, we initially prompt all baseline models as well as Mistral-Interact to explicitly judge task vagueness, ask for missing details, and summarize the user's goal.

\subsubsection{Metrics}
\paragraph{Main Metrics.} We present novel metrics that convert subjective human intentions in user-agent interactions into objective numerical values, thus simplifying data analysis and comparison.
\begin{itemize}[topsep=0pt, partopsep=1pt, leftmargin=12pt, itemsep=0pt]
\item \textbf{Vagueness Judgement Accuracy}: We calculate the percent of the model's judgments of task $t$'s vagueness (vague or clear) that are aligned with ground truth. This measures the model's ability to discern vagueness from clearance and avoid inquiring about already-clear tasks.
\item \textbf{Missing Details Recover Rate}: For ground truth missing details of different importance levels, we analyze what percent are recovered (explicitly inquired) by the model during the interaction. This measures the model's ability to prioritize inquiring about necessary details.
\item \textbf{Summary Intention Coverage Rate}: The percent of user-provided intentions that are explicitly summarized finally in $t_{user}$ by the model. This measures the model's ability to summarize user intentions comprehensively and explicitly without repetitions or omissions.
\end{itemize}

\paragraph{Other Metrics.} Despite the three main metrics that directly reflect the model's ability to understand the user's implicit intentions, we also analyze other conversation details for a more comprehensive evaluation.
\begin{itemize}[topsep=0pt, partopsep=1pt, leftmargin=12pt, itemsep=0pt]
\item \textbf{Options Presenting Rate}: For all the missing details explicitly queried by the model, we analyze the percent of them accompanied by potential options.
\item \textbf{Options Reasonable Rate}: For options provided by the model, we record the percent of them that the user believes is reasonable to propose. This measures whether the model could \emph{actively and positively} inspire user responses.
\item \textbf{Average Provided Options}: Average number of options the model provides for one missing detail during the inquiry. 
\item \textbf{Average Inquired Missing Details}: Average number of missing details the model inquires for one task.
\item \textbf{Average Conversation Rounds}: Average number of conversation rounds that the model has with the user for one task.
\item \textbf{Average Inquired Missing Details Per Round}: Average number of missing details the model inquires for one round of conversation. 
\end{itemize}
We provide the formalized definition and detailed calculation formulas for each metric in \Cref{apdx:interact_exp_metric_calculation}.

\paragraph{Measurement Details.} For all the metrics, we apply direct statistical calculation, user annotation during conversations, or involvement of GPT-4 to help with our measurement, detailed in \Cref{apdx:interact_exp_metric_measurement}. In addition, we provide only the macro-average calculation results over all the testing tasks, as the micro-average results reflect approximately the same trend among models.

\begin{table*}[!t]
\begin{center}
\footnotesize
\tabcolsep=0.018\linewidth
\begin{tabular}{llrrrr}
\toprule
\multicolumn{2}{l}{\textbf{Metrics}} & \textbf{Mistral-7B} & \textbf{LLaMA-2-7B} & \textbf{GPT-4} & \textbf{Mistral-Interact} \\
\midrule
\multicolumn{2}{l}{\textbf{$^{\uparrow}$Vagueness Judgement Accuracy (\%)}}                     &    49.07    &    79.63    &   82.41    &       \textbf{85.19}           \\
\multirow{3}{*}{\textbf{\makecell{Missing Details Recover Rate \quad\quad\\ (of Importance Level, \%)}}}
& \textit{- $^{\uparrow}$Lv 3} &      68.42      &       60.98     &    \textbf{75.22}   &      72.28            \\
& \textit{- $^{\uparrow}$Lv 2} &      56.94      &       38.76     &    63.14   &      \textbf{67.08}            \\
& \textit{- \ \ Lv 1} &      23.08      &       28.92     &    37.50   &      27.94            \\
\multicolumn{2}{l}{\textbf{$^{\uparrow}$Summary Intention Coverage Rate (\%)}}                     &      91.43      &     61.87       &    \textbf{100.0}   &       96.37           \\
\midrule
\multicolumn{2}{l}{\textbf{$^{\uparrow}$Options Presenting Rate (\%)}}                             &       42.46     &      47.64      &    40.31   &       \textbf{84.08}           \\
\multicolumn{2}{l}{\textbf{$^{\uparrow}$Options Reasonable Rate (\%)}}                             &      \textbf{100.0}      &        81.79    &    \textbf{100.0}   &       98.70           \\
\multicolumn{2}{l}{\textbf{Average Provided Options}}                            &      1.46      &       1.35     &   1.21    &       2.72           \\
\multicolumn{2}{l}{\textbf{Average Inquired Missing Details}}                    &      3.91      &       5.80     &   4.78    &       4.52           \\
\multicolumn{2}{l}{\textbf{Average Conversation Rounds}}                         &      1.62      &     3.02       &    2.69   &         4.15         \\
\multicolumn{2}{l}{\textbf{$^{\downarrow}$Average Inquired Missing Details Per Round}}          &      2.80      &       2.49     &    2.31   &        \textbf{1.26}          \\
\bottomrule
\end{tabular}
\end{center}
\caption{The results of Mistral-Interact and baselines regarding different metrics on the test split tasks of IN3. Arrows represent the higher ($\uparrow$) or the lower ($\downarrow$) the better.}
\label{tab:main_results}
\end{table*}

\subsubsection{Results}

From the overall results presented in \Cref{tab:main_results}, we summarize our findings about Mistral-Interact as follows. 

\paragraph{Better understanding of user judgments.}
Among all the open-source models, Mistral-Interact is the best at predicting task vagueness and missing details that users regard as necessary (especially \textit{Lv 3} and \textit{Lv 2}). We show that Mistral-Interact's vagueness judgment is the most accurate, and it could recover over 70\% of the most important missing details, far better than LLaMA-2-7B and Mistral-7B while rivaling the performance of GPT-4. This can be attributed to the structured and comprehensive initial thoughts in constructed conversation records applied to Mistral-Interact's training.

\paragraph{Comprehensive summarization of user intentions.} 
Mistral-Interact is effective in making an explicit and comprehensive summary based on detailed user intentions. We observe that Mistral-Interact, compared with other open-source models, has the highest average conversation rounds around 4.5. Despite the resulting more user-provided information, it still gives the most comprehensive summaries with fewer omissions, covering over 96\% of all user intentions, which is challenging to achieve.

\paragraph{Enhanced model-user interaction experience.} 
Mistral-Interact inquires about missing details in vague tasks more reasonably and friendly than other open-source models, thus promoting a clearer understanding of the user's implicit intentions. This entails Mistral-Interact asking fewer questions per round (approximately only one) but still maintaining a high recover rate and providing a multitude of reasonable options for most missing details. These traits better inspire the users to reveal their inner worlds while making them more willing to respond instead of feeling stuffed with inquiries.

\paragraph{Comparable performance with closed-source GPT-4.} 
We prove that smaller-scale model experts can approach or even exceed general-purpose large-scale models across various aspects including vagueness judgment, comprehensiveness of summaries, and friendliness of interaction. According to the presented results, Mistral-Interact's performance closely matches that of GPT-4 across all metrics, even surpassing GPT-4's capability on vagueness judgment, option provision, and coverage of certain missing details. In addition, compared with GPT-4, Mistral-Interact is also more cost-efficient while achieving comparable performance.

\begin{figure*}[!t]
    \centering
    \subfigure{\includegraphics[width=\linewidth]{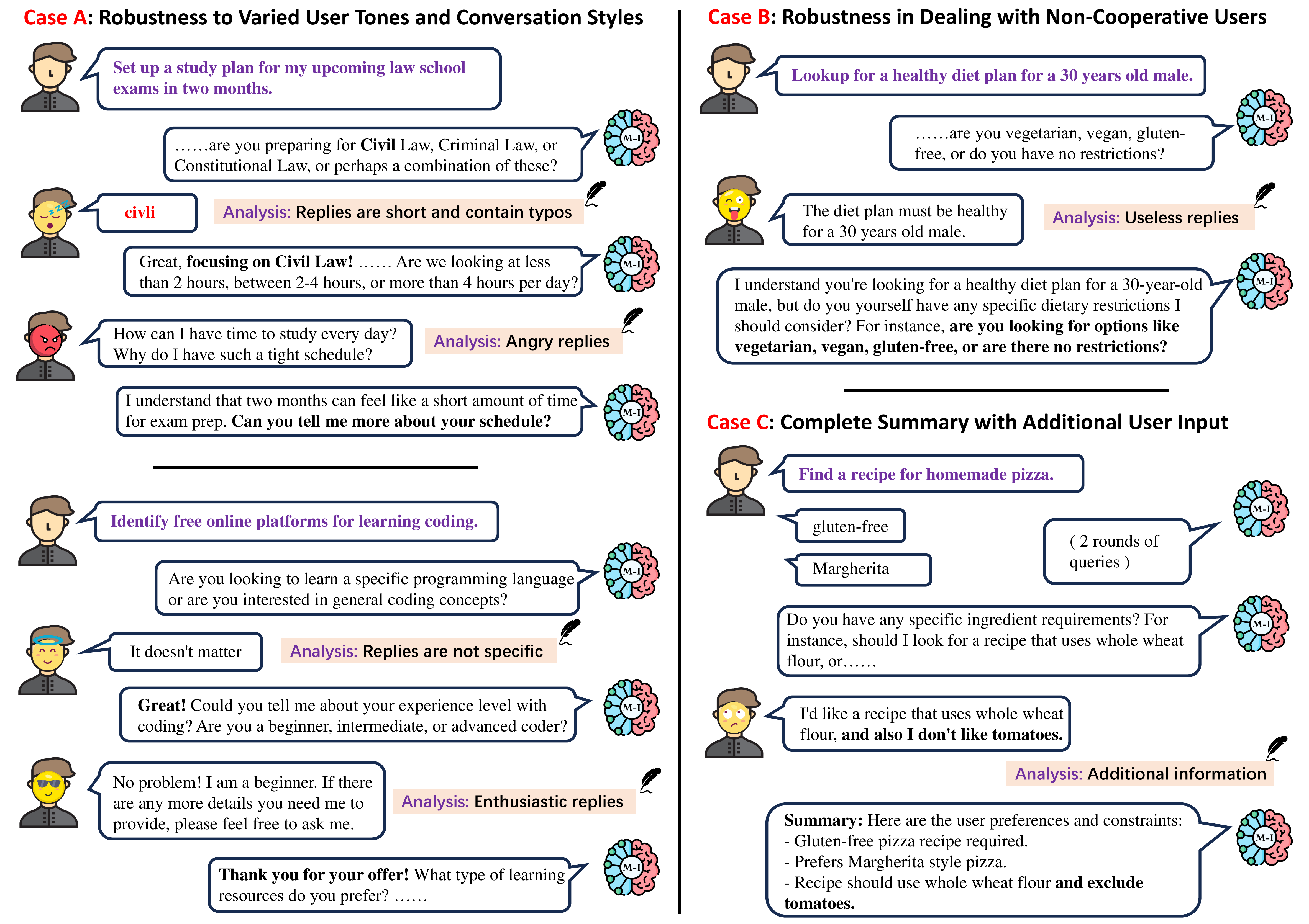}}
    \caption{Case studies of model-user interactions under different scenarios to show Mistral-Interact's robustness.}
    \label{fig:case_study}
\end{figure*}

\subsubsection{Case Study}
To further show Mistral-Interact's robustness under different conversation scenarios, we present three case studies in \Cref{fig:case_study} and illustrate respectively as follows.

\paragraph{Robustness to varied user tones and conversation styles.} In Case A of \Cref{fig:case_study}, we show the impact of varied user tones and conversation styles on Mistral-Interact's responses. We discover that regardless of whether the user's responses are brief or detailed, enthusiastic or aloof, and even if containing typos, Mistral-Interact can comprehend them accurately and provide appropriate responses, which proves its robustness.

\paragraph{Robustness in dealing with non-cooperative users.} In Case B of \Cref{fig:case_study}, we examine whether Mistral-Interact can continue probing and steer the conversation back on track when users display reluctance or non-cooperation. The result shows that even when users evade the question, our adapted model can still redirect the conversation effectively.

\paragraph{Additional User Information.} In Case C of \Cref{fig:case_study}, we observe that Mistral-Interact can incorporate additional user-provided information that is not explicitly inquired by the assistant into the summary. This indicates that when the model does not fully cover missing details or the user has specific requirements, the model can still reasonably and comprehensively summarize all user intentions, thus making it more user-friendly.

\subsection{Evaluation on Instruction Execution}
To evaluate the effectiveness of the implicit intention understanding for instruction execution, we integrate Mistral-Interact as an upstream interaction module into the XAgent framework \citep{xagent2023}, an autonomous agent system for complex task solving.
We conduct proof-of-concept experiments through performance comparison and case studies, as illustrated in this section.

\subsubsection{Experimental Settings}
\paragraph{Data.} We randomly sample ten tasks from the IN3's test split that are i) regarded as vague by both the ground truth and Mistral-Interact, and ii) within the XAgent's ability. For these tasks, we extract the initial vague task description and the summary with clear user intentions derived from the user-agent interaction. They are then both sent for agent execution for evaluation and comparison. Each task's detailed content is presented in \Cref{apdx:agent_exp_data}.

\paragraph{Settings.} We enable the plan and execution engine of the XAgent framework. The environments where XAgent can be grounded for interaction include web searching, code execution, command line, and file system. We employ GPT-4 as the default model for both task planning and execution.

\subsubsection{Metrics}
The XAgent divides the user's task into subtasks during planning, each with a set of milestones that the agent aims to achieve through tool invocation chains. We design metrics that aim to reflect the effectiveness of our method in clarifying the user's task goal and making the agent execution more efficient. Specifically, we measure:
\begin{itemize}[topsep=0pt, partopsep=1pt, leftmargin=12pt, itemsep=0pt]
\item \textbf{Unnecessary Subtasks / Milestones}: The percent of subtasks or milestones that are regarded as unnecessary by the user under the detailed task goal with clear user intentions.
\item \textbf{General Subtasks / Milestones}: The percent of subtasks or milestones that are too general, instead of focusing on the user's specific intentions.
\item \textbf{Tool Invocations Per Subtask / Milestone}: The average number of tool invocations for one subtask or milestone, which reflects the efficiency of agent execution.
\end{itemize}
Please refer to \Cref{apdx:agent_exp_data} for details on the measurement of all metrics.

\subsubsection{Results}

\begin{table*}[!t]
\centering
\footnotesize
\tabcolsep=0.03\linewidth
\begin{tabular}{cccc}
\toprule
\textbf{Scenarios} & \textbf{\makecell{Unnecessary \\ ST / MS (\%)}} & \textbf{\makecell{General \\ ST / MS (\%)}} & \textbf{\makecell{Tool Invocations Per\\ ST / MS}} \\
\midrule
\textbf{Vague Task} (w/o Mistral-Interact) & 22.22 / 21.48 & 22.22 / 12.08 & 5.22 / 2.21  \\
\textbf{Detailed Task} (w/ Mistral-Interact) & \textbf{1.85} / \textbf{7.81} & \textbf{0.00} / \textbf{0.78} & \textbf{4.79} / \textbf{2.02} \\
\bottomrule
\end{tabular}
\caption{The comparison of agent task performance with or without the incorporation of Mistral-Interact as an upstream module for interaction. \emph{ST} denotes subtask and \emph{MS} denotes milestone.}
\label{tab:exp_xagent}
\end{table*}

We present the quantitative evaluation results in \Cref{tab:exp_xagent} and discover that our method is helpful to i) avoid setting unnecessary goals during execution, ii) make the agent more aligned with detailed user intentions, and iii) facilitate agent tool execution efficiency. All these aspects reflect Mistral-Interact's effectiveness as an upstream user-agent interaction expert in promoting downstream agent execution.

\subsubsection{Case Study}

\begin{figure*}[!t]
    \centering
    \subfigure{\includegraphics[width=\linewidth]{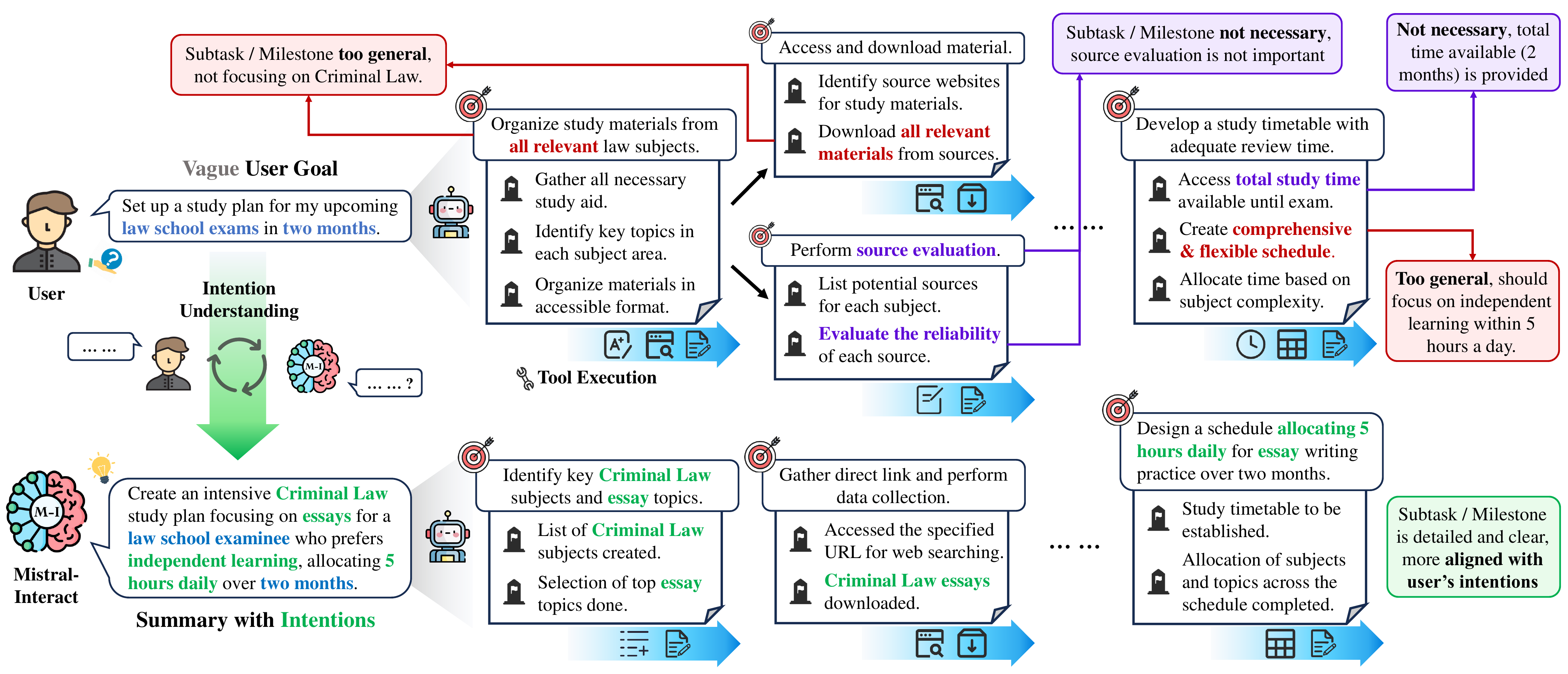}}
    \caption{Case study on the agent execution process before and after interaction with Mistral-Interact in agent design.}
    \label{fig:xagent_case_study}
\end{figure*}

To clearly illustrate Mistral-Interact's role, we present a case study through comparison in \Cref{fig:xagent_case_study}. From the phrases marked red, we reveal that when the user's goal is vague, XAgent tends to set general subtasks and milestones, instead of focusing on one specific law subject or taking into account the user's actual available time. From the phrases marked purple, we reveal that XAgent also sets subtasks and milestones that are sometimes unnecessary. This occurs as the user's task is too vague to execute and XAgent tends to make up unnecessary details (e.g. source evaluation), which does not align with the user's true intentions.

In contrast, after active interaction, the clear task goal promotes tailored and specific subtasks and milestones. We show this alignment in phrases marked in green. At the same time, the execution flow becomes much more simplified, and the tools applied become more focused, with tool invocation times significantly decreasing. All of these reflect a more effective and efficient agent execution process.

%% file: Sections/5_Discussions.tex
\section{Discussions}

\paragraph{Incorporating model-user interaction in agent scenarios.}
Currently, agent designs typically relegate humans to peripheral roles, which motivates our work on improving the language model's ability to understand user's implicit intentions before agent execution. This limitation can also be addressed through alternative methods, including mechanisms that facilitate interactions during agent execution. Ideally, an agent system should be able to promptly query users when faced with multiple options, unclear instructions, critical tool invocation errors, etc. Moreover, users ought to be more actively involved in the agent execution, which could involve arbitrary user interruptions throughout the tool invocation process, the granting of permissions for potentially risky agent actions, etc. Both aspects require complex and nuanced coordination within the agent system, as opposed to simply adapting and leveraging a single model expert's capability. This remains an exciting field for future research efforts.

\paragraph{Assessing model-user interaction.}
Our primary method for evaluating interaction is through implicit intention understanding, both directly via the adapted model's performance and indirectly through the downstream agent's execution. The metrics we propose could be improved to encompass additional facets. For instance, during agent execution, user inputs are unpredictable, and thus the interpretation of user intentions could include more dimensions like whether they are providing more information, interrupting an execution, asking for progress updates, or introducing a new topic, rather than solely focusing on task vagueness. Furthermore, while we strive to quantify subjective human evaluations into objective numerical values for comparison, other techniques could also be integrated, such as direct evaluation of user satisfaction with the interaction, their immediate perceptions of the conversation's coherence, and their opinions on the downstream agent's final outputs. These could yield a more holistic evaluation, albeit potentially introducing individual bias.

\paragraph{Employing language models to emulate users.}
In constructing the IN3 dataset, we used GPT-4 to imitate users, leveraging the model's proficiency in emulating different roles via careful prompting. For example, it can effectively imitate users with varying tones (e.g., angry, passionate) and response styles (e.g., succinct, verbose). This possesses broad implications as it: i) encourages unsupervised model-(model-simulated) user interactions, which could subsequently be used for the assistant model's reinforcement learning; ii) supports automated agent evaluation, as benchmarks could be created using solely model-simulated users, bypassing the need for real users in time-consuming agent executions and interactions. In addition, the current method is agnostic of real user preference data; if the model can access an individual's past conversation history, it could better represent specific individual preferences, thus facilitating better coherency and faithfulness of implicit user intentions.

%% file: Sections/6_Conclusion.tex
\section{Conclusion}
This work investigates the enhancement and evaluation of implicit intention understanding in agent designs. Specifically, we release the Intention-in-Interaction (IN3) benchmark which encompasses a wide range of vague agent tasks with annotated missing details to evaluate the agent's intention understanding capability. In addition, we propose to integrate a special model expert into agent designs for robust interaction. Empirically, we train Mistral-Interact, a powerful open-source model capable of effectively judging user task vagueness, friendly inquiring about missing details with reasonable options, and comprehensively summarizing user intentions without omission. Integrating Mistral-Interact into the XAgent framework, we evaluate the enhanced agent system through both instruction understanding and instruction execution to illustrate our method's effectiveness.
Overall, we are among the first works to focus on user participation and implicit intention understanding in agent designs and evaluations. We hope our work can inspire more future research on robust human-agent interaction evaluation benchmarks and mechanisms that enhance efficiency and effectiveness.

%% file: Sections/Appendix.tex
\clearpage
\appendix

\section*{Appendix}
\label{sec:appendix}

\section{Intention-in-Interaction Benchmark}

\subsection{Task Generation}
\label{apdx:in3_task}
For the tasks in IN3, we manually create seed data and generate agent tasks iteratively using the system prompt as follows. For each category, we calculate the embeddings of the task using text-embedding-ada-002 and perform filtering to ensure the cosine similarity between any two tasks is less than 0.8.\\
\makebox[\linewidth]{\rule{\linewidth}{0.4pt}}
\textit{Instruction (System Prompt)}
\begin{lstlisting}[basicstyle=\small\ttfamily, breaklines=true, breakindent=0em]
You are a task-generation engine. Your mission is to generate tasks in everyday life that could be fulfilled by an agent. The agent working for you has the following accesses:
--- Agent Resources ---
- Internet Access for searches and information gathering. 
- A File System Environment to read and write files (text, code, markdown, latex...). 
- A Python Notebook to execute Python code.
- A ShellEnv with root privilege to execute bash command.
--- Task Description ---
Based on what you know about the agent, you can generate tasks that are suitable for the agent to solve. You should generate tasks in a first-person tone, it should be clear, but don't provide too many details or unnecessary information.
--- Important Note ---
- Make your generated tasks as diverse as possible. The user will provide some examples but do not copy the contents.
- Generate tasks of different difficulties, and they are all solvable by the agent using resources.
- The tasks, should be grounded in the real world, but also keep it vague. Use just one sentence and do not provide too many details.
- Please list your generated tasks through tool call.
\end{lstlisting}

\subsection{Human Annotation}
\label{apdx:in3_annotation}
During annotation, the importance level of each missing detail is annotated through the following rules. Lv 3: Very important, task cannot be fully executed without it; Lv 2: Relatively important: knowing it can better help the user execute the task, but not that necessary; Lv 1: Not very important, it is too detailed or general, the task can still run successfully without knowing it. Note that if as task is annotated as clear, its further annotations on missing details, importance scores and potential options will all be omitted.

We employ different people with diverse backgrounds to annotate the data, each person in charge of several tasks. The annotations could thus reflect different people's preferences and ensure diversity. We also employ GPT-4 to help with the annotation. GPT-4 will provide suggestions on task vagueness and potentially missing details as references, and annotators only need to do the filtering and information updates to make annotations aligned with their preferences and intentions. GPT-4's suggestions are completed through tool calling, with the system prompt and the specific tool as follows:\\
\makebox[\linewidth]{\rule{\linewidth}{0.4pt}}
\textit{Instruction (System Prompt)}
\begin{lstlisting}[basicstyle=\small\ttfamily, breaklines=true, breakindent=0em]
You are an agent judging if the user's task goal is vague or not.
Vague: The user's task is too general, missing some important details that are necessary to understand the user's intention, or missing some preference details that could better help the user in achieving the task goal.
Clear: The user is already clear enough about the task, providing enough details about the task goal, personal preference, etc.
  
If the task is vague, provide what details are missing, or what further information is needed. There may be multiple missing details.
For each missing information, please also provide a query to the user asking for this missing information, and provide a list of options that the user could choose from.
\end{lstlisting}
\makebox[\linewidth]{\rule{\linewidth}{0.4pt}}
\textit{Function Tool}
\begin{lstlisting}[basicstyle=\small\ttfamily, breaklines=true, breakindent=0em]
name: judge_vagueness
  description: "Judge if the user's task goal is vague or not, and provide what details or personal preferences are missing."
  parameters:
    type: "object"
    properties:
      thought:
        type: "string"
        description: "Generate thought about why this task goal is vague or clear. Please refer to the description of vague and clear in the system prompt."
      judgment:
        type: "string"
        enum: ["vague", "clear"]
        description: "Based on your thought, choose if the task is vague or clear."
      missings:
        type: "array"
        description: "The details that are missing in understanding the user's task intention or preferences. Leave the array empty if the task is clear. There could be multiple missing details. Each missing detail should be specific."
        items:
          type: "object"
          properties:
            description:
              type: "string"
              description: "Description of what detail is missing in a short way."
            importance:
              type: "string"
              enum: ["3", "2", "1"]
              description: "Give each missing detail an importance score. 3: Very important, task cannot be fully executed without it; 2: Relatively important: knowing it can better help the user execute the task, but not that necessary; 1: Not very important, it is too detailed or general, the task can still run successfully without knowing it."
            inquiry:
              type: "string"
              description: "Form an inquiry to the user asking for this missing detail. Inquiry like you are chatting with the user."
            options:
              type: "array"
              description: "Provide some possible options for the user to choose from. Please always provide 2-3 possible options or suggestions to inspire the user."
              items:
                type: "string"
                description: "Options or suggestions to this missing detail. Make the options very short and specific (e.g. just using phrases)"
          required: ["description", "inquiry", "importance", "options"]
    required: ["thought", "judgment", "missings"]
\end{lstlisting}

\section{Preliminary Experiment}

\subsection{Settings}
\label{apdx:preliminary_exp}

\paragraph{Tasks.} We randomly sampled ten tasks from IN3 to conduct a preliminary test on different models' intention understanding ability. The specific tasks are as follows:\\
\makebox[\linewidth]{\rule{\linewidth}{0.4pt}}
\textit{Tasks List}
\begin{lstlisting}[basicstyle=\small\ttfamily, breaklines=true, breakindent=0em]
1. Category: Health Research
Task: Find the latest research on diabetes treatment.
Vague: True
Missing Details:
- Type of diabetes
- Aspect of treatment
- Source of research
- time of when the research is published

2. Category: Animal Studies
Task: Find the average lifespan of a domestic cat.
Vague: False
Missing Details: None

3. Category: Coding Project
Task: Write a Python script that sorts images based on their content.
Vague: True
Missing Details:
- Criteria for sorting images
- Source of images
- Output format
- Preferred libraries or methods
- Level of accuracy expected

4. Category: Financial
Task: Look up the current exchange rate from USD to Euro.
Vague: False
Missing Details: None

5. Category: Writing Task
Task: Write a summary of a given text file.
Vague: True
Missing Details:
- Purpose of the summary
- Subject matter of the text file
- Intended audience
- Path to the text file
- Length of the summary

6. Category: Traveling
Task: Find the cheapest round-trip flights from New York to London next month.
Vague: True
Missing Details:
- Specific dates of travel within the month
- Preferred airlines, airports, or alliances
- Preferred time of day for flying
- Cabin class preference

7. Category: Hardware
Task: Write bash command to check my system specification.
Vague: True
Missing Details:
- Specific system specifications of interest
- Output format

8. Category: Health Advise
Task: Locate the nearest yoga class with the best reviews in my city.
Vague: True
Missing Details:
- User's city
- Distance willing to travel
- Criteria for best reviews
- Preferred style of yoga

9. Category: Music Analysis
Task: Find out what the top five Billboard hits are this week.
Vague: False
Missing Details: None

10. Category: Anime
Task: Find a list of the top-rated Anime series in 2021.
Vague: False
Missing Details: None
\end{lstlisting}

\paragraph{Model Setting.} For the specific models, we fully considered their conversation ability and applied the newest versions. We finally chose Mistral-7B-Instruct-v0.2, LLaMA-2-7B-Chat, and GPT-4-1106 as the testing model. For all the models, we set the temperature to 0.7, top\_p to 0.9, and maximum sequence length to 2048. Each task is tested once.

\paragraph{System Prompt.} We apply the same instruction to prompt all three models. The detailed content is as follows:\\
\makebox[\linewidth]{\rule{\linewidth}{0.4pt}}
\textit{Instruction (System Prompt)}
\begin{lstlisting}[basicstyle=\small\ttfamily, breaklines=true, breakindent=0em]
You are trying to understand the user's intention. The user will provide a task that may be clear or vague (they may not provide their own preferences).
1. In the very first round of thought, you should explicitly judge if the task is vague or clear and why.
2. If the task is vague, you should ask the user for more information with options for the user to choose from. If it is clear, then do not query and repeat the user's task in the summary.
3. Please only ask one question with options at a time. You could ask the user for multiple rounds until you think the user's goal is clear enough.
4. Your questions should be about different missing details and aspects, be diverse.
5. When you think you have gathered enough information, you should provide a summary of the user's detailed goal.
6. Do not solve the task, just provide a detailed summary of the task and user preference in the end. Summarize the information you got in 1-2 sentences.

You should follow the format, first provide your thoughts, then ask the user for more information or provide a summary: 
Thought: ... Query: ... (You would like to query the user)
OR: Thought: ... Summary: ... (You have gathered enough information and could make a summary)
The user will provide a task as the following.
\end{lstlisting}

\subsection{Evaluation}
\label{apdx:preliminary_eval}
We conduct both quantitative and qualitative evaluations of the model performance. As the number of tasks is small, the quantitative results are counted manually by comparing the model's response with the annotation.

For the qualitative analysis, we manually identify the failure patterns in the interaction records. For each pattern, we consider it as \deepred{\textbf{severe}} if a certain model exhibits this pattern in more than three tasks among ten, and consider it as \lightred{\textbf{light}} if a certain model exhibits this pattern but in less than (or equal to) three tasks. All the judgments are consensus among at least three different annotators.

\subsection{Case Study}
\label{apdx:preliminary_case}

\begin{figure*}[!t]
    \centering
    \subfigure{\includegraphics[width=\textwidth]{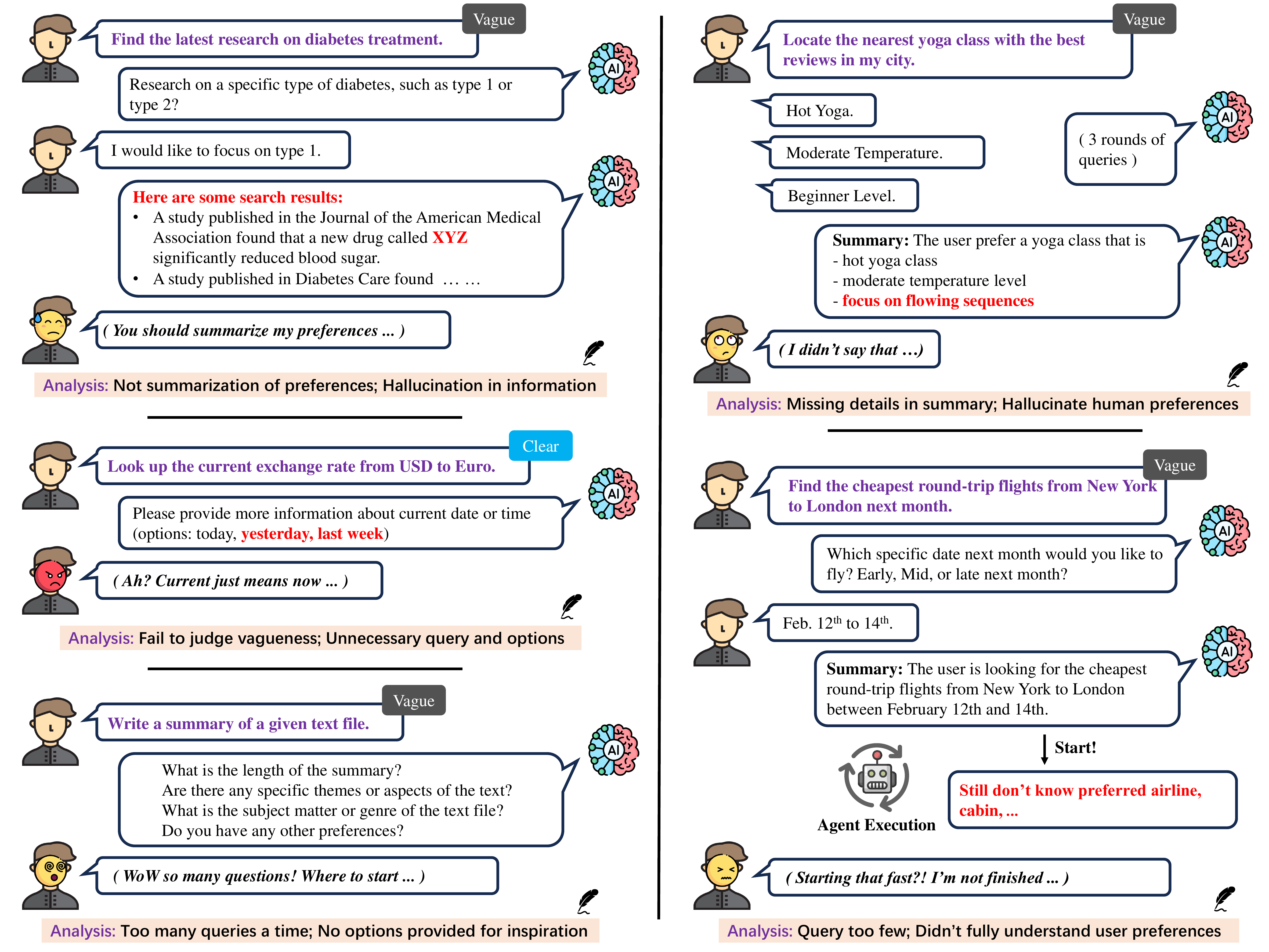}}
    \caption{An overview of different failed model-user interaction patterns.}
    \label{fig:preliminary}
\end{figure*}

We present a case study on different failed interaction patterns between the model and user in \Cref{fig:preliminary}. All the cases are gathered from real model-user conversation records with analysis on the specific inappropriate behaviors. For either vague or clear tasks, we show how and why the model responses in interactions make the user unsatisfied.

\section{Construction of Training Data}
\label{apdx:dataset_conversation}
For each task in IN3, we apply GPT-4 to simulate both the assistant (Assistant-GPT) and the user (User-GPT) during the conversation. The conversation consists of three parts, respectively initial thought, inquiries, and summary. 

\paragraph{Initial Thought.} According to the annotation for each task, we manually construct the explicit initial thought by concatenating information on the judgment of task vagueness, the reasons for it, the missing details if exists, and the options for each missing detail.

\paragraph{Inquiries.} Next, we prompt both Assistant-GPT and User-GPT to simulate the conversation with the following instructions. Note that to simulate different user tones, we create two versions of instructions.\\
\makebox[\linewidth]{\rule{\linewidth}{0.4pt}}
\textit{Assistant Side Instruction (System Prompt)}
\begin{lstlisting}[basicstyle=\small\ttfamily, breaklines=true, breakindent=0em]
You are an agent trying to specify and understand the user's task goal. The user will ask you a query or ask you to execute a task. However, the user is unclear about the task or intention. You should ask the user for more information to understand the user's intention.

Here are some rules to follow:
1. You are given the initial thought and a list of possible inquiry aspects and an option list. Please use this information as a reference when inquiring.
2. For each inquiry, provide the user with options or some suggestions. Use a first-person tone like chatting with the user, and be friendly.
3. You can ask either a new question (from the reference, with options) or a follow-up inquiry (from the user's last response). Please use thought to show why you made this inquiry.
4. Please only inquire for one question in one round of chatting. You can inquire for multiple rounds, but please control the total rounds to be less than five. (The user is impatient, make your inquiry efficient!)
5. Choose to stop if you think the information you have gathered is enough. Remember you don't need to ask for every detail!

You are talking about <category> with the user. This is what you'd like to ask or do: <task>
This is your initial thought: <thought>
This is the list of possible inquiry aspects (reference list):
<missing details>
\end{lstlisting}
\makebox[\linewidth]{\rule{\linewidth}{0.4pt}}
\textit{User Side Instruction (System Prompt)}
\begin{lstlisting}[basicstyle=\small\ttfamily, breaklines=true, breakindent=0em]
You are an assistant who pretends to be the user's friend and responds to the user. The user is trying to understand your specific needs and intentions and may ask you some questions. You should provide the information to the user in one sentence.

Here are some tips during chatting to make your response more real.
[Passionate User Tone Version]
1. Respond naturally, and you are passionate. You can provide more if you are happy with it. Keep your tone friendly and positive.
[Succinct User Tone Version]
1. Respond succinctly, and you are lazy. You should respond more often with short phrases. Make your responses short and effective.

2. When you are asked about some personal preference, information, or address, please make up some information and preference and provide it to the user. Make sure to be specific and as real as possible.

You are talking about <category> with the user. This is what you'd like to ask or do: <task>
\end{lstlisting}

\paragraph{Summary.} We prompt Assistant-GPT to make explicit summary thoughts and final summary through tool call. Specifically, the summary thought is manually constructed by explicitly concatenating the original thought and a list of constraints that reflect user preferences or new additional information. If the original task is clear, then the constraints are omitted. The instruction and tool are respectively as follows:\\
\makebox[\linewidth]{\rule{\linewidth}{0.4pt}}
\textit{Instruction (System Prompt)}
\begin{lstlisting}[basicstyle=\small\ttfamily, breaklines=true, breakindent=0em]
You are an agent trying to summarize the user's intention and provide a detailed summary.

First, provide thought about why you think you have gathered enough information to understand the user's intention, or why the initial task is clear enough.
Secondly, if there is an interaction history, explicitly list the user's provided constraints or preferences one by one in a list.
Lastly, provide a detailed summary, including the task goal and all the user's constraints and preferences. You should respond naturally within 2 sentences (make your language succinct, short, and efficient).

The user's original task is: <task>
\end{lstlisting}
\makebox[\linewidth]{\rule{\linewidth}{0.4pt}}
\textit{Function Tool}
\begin{lstlisting}[basicstyle=\small\ttfamily, breaklines=true, breakindent=0em]
name: complete_summary
  description: "Complete the summary by providing thought, listing user preferences and constraints, and providing a detailed summary. Respond naturally and succinctly."
  parameters:
    type: "object"
    properties:
      thought:
        type: "string"
        description: "Why do you think you have gathered enough information to understand the user's intention, or why the initial task is clear enough."
      constraints:
        type: "array"
        description: "A list of user preferences and constraints based on the interaction history. The number of items should be equal to the rounds of chatting. Leave the array empty if the initial task is clear."
        items:
          type: "string"
          description: "The user's preference or constraint in the first, second, third, etc. round of chatting. Summarize and list them one by one. Make it detailed and succinct."
      summary:
        type: "string"
        description: "Summarize the user's task goal and the constraints in a detailed, efficient, and succinct way within two sentences. Do not provide not-mentioned or unnecessary information."
    required: ["thought", "constraints", "summary"]
\end{lstlisting}

\section{Training Details}
\label{Apdx:training_details}

\subsection{System Prompt}
We apply the following system prompt to train the model. It is incorporated at the beginning of each final training data point. The detailed content is as follows:\\
\makebox[\linewidth]{\rule{\linewidth}{0.4pt}}
\textit{Instruction (System Prompt)}
\begin{lstlisting}[basicstyle=\small\ttfamily, breaklines=true, breakindent=0em]
You are an agent trying to understand the user's goal and summarize it. Please first ask users for more specific details with options, and finally summarize the user's intention.
--- Step 1: initial thought generation ---
1. Generate [INITIAL THOUGHT] about if the task is vague or clear and why.
2. List the important missing details and some according options if the task is vague.
--- Step 2: inquiry for more information if vague ---
1. If the task is vague, inquire about more details with options according to the list in [INITIAL THOUGHT].
2. Think about what information you have and what to inquire next in [INQUIRY THOUGHT].
3. Present your inquiry with options for the user to choose after [INQUIRY], and be friendly.
4. You could repeat Step 2 multiple times (but less than 5 times), or directly skip Step 2 if the user task is clear initially.
--- Step 3: summarize the user's intention ---
1. Make the summary once the information is enough. You do not need to inquire about every missing detail in [INITIAL THOUGHT].
2. List all the user's preferences and constraints in [SUMMARY THOUGHT]. The number of points should be the same as rounds of chatting.
3. Give the final summary after [SUMMARY] with comprehensive details in one or two sentences.
\end{lstlisting}

\subsection{Conversation Prompt}
We apply the following conversation template to convert and concatenate each conversation record for training. For each constructed conversation record for training, we directly apply all the User-GPT responses. However, the Assistant-GPT's responses include thoughts, judgments, inquiries, and summaries, so we involve three steps in the template:
\begin{itemize}[topsep=0pt, partopsep=1pt, leftmargin=12pt, itemsep=0pt]
\item \textbf{First round}: The model needs to tell the vagueness of the task. If it is vague, the model should provide the missing details, and then proceed with the first-round inquiry.
\item \textbf{Middle rounds}: If the task is vague, the model should continue asking the user for any missing details.
\item \textbf{Last round}: After the model has gathered sufficient information (or the task is already clear), it should summarize the user's intention and detailed task goal with thoughts.
\end{itemize}

Based on whether the task is vague or clear, the data point should fit into different templates as follows:\\
\makebox[\linewidth]{\rule{\linewidth}{0.4pt}}
\textit{Conversation Template (Vague Task)}
\begin{lstlisting}[basicstyle=\small\ttfamily, breaklines=true, breakindent=0em]
{System Prompt}

### First round
[INITIAL THOUGHT] {initial_thought} Some aspects of missing details and potential options are as follows:
{missing details}
[INQUIRY THOUGHT] {inquiry_thought}
[INQUIRY] {inquiry}
(User Response)

### Middle rounds (* N)
[INQUIRY THOUGHT] {inquiry_thought}
[INQUIRY] {inquiry}
(User Response)

### Last round
[SUMMARY THOUGHT] {summary_thought}
[SUMMARY] {summary}
\end{lstlisting}
\makebox[\linewidth]{\rule{\linewidth}{0.4pt}}
\textit{Conversation Template (Clear Task)}
\begin{lstlisting}[basicstyle=\small\ttfamily, breaklines=true, breakindent=0em]
{System Prompt}

### First round
[INITIAL THOUGHT] {initial_thought}

### Last round
[SUMMARY THOUGHT] {summary_thought}
[SUMMARY] {summary}
\end{lstlisting}

\subsection{Complete Training Sample Example after Concatenation}
We apply the aforementioned system prompt and conversation template to transform all the conversation records for training. An example of complete training data after this process is as follows:\\
\makebox[\linewidth]{\rule{\linewidth}{0.4pt}}
\textit{Example}
\begin{lstlisting}[basicstyle=\small\ttfamily, breaklines=true, breakindent=0em]
<s> User: You are an agent trying to understand the user's goal and summarize it. Please first ask users for more specific details with options, and finally summarize the user's intention.
--- Step 1: initial thought generation ---
1. Generate [INITIAL THOUGHT] about if the task is vague or clear and why.
2. List the important missing details and some according options if the task is vague.
--- Step 2: inquiry for more information if vague ---
1. If the task is vague, inquire about more details with options according to the list in [INITIAL THOUGHT].
2. Think about what information you have and what to inquire next in [INQUIRY THOUGHT].
3. Present your inquiry with options for the user to choose after [INQUIRY], and be friendly.
4. You could repeat Step 2 multiple times (but less than 5 times), or directly skip Step 2 if the user task is clear initially.
--- Step 3: summarize the user's intention ---
1. Make the summary once the information is enough. You do not need to inquire about every missing detail in [INITIAL THOUGHT].
2. List all the user's preferences and constraints in [SUMMARY THOUGHT]. The number of points should be the same as rounds of chatting.
3. Give the final summary after [SUMMARY] with comprehensive details in one or two sentences.

Here is the task:
I would like to plan a trip to Paris next month.

Agent:  [INITIAL THOUGHT] The user's task of planning a trip to Paris next month is vague because it lacks specific details such as the duration of the trip, the budget, preferences for accommodation, activities they are interested in, and whether they need help with transportation or dining options. Some aspects of missing details and potential options are as follows:
- Duration of the trip: 3-5 days, 1 week, More than a week
- Budget for the trip: Economy, Mid-range, Luxury
- Accommodation preferences: Hotel, Hostel, Apartment rental
- Activities of interest: Sightseeing, Cultural experiences, Shopping
- Transportation needs: Public transport information, Car rental, Private driver
- Dining preferences: Local cuisine, Vegetarian/Vegan options, No preferences
[INQUIRY THOUGHT] The user wants to plan a trip but hasn't specified the duration, which is essential to start planning.
[INQUIRY] How exciting! How long are you planning to stay in Paris? Would it be something like 3-5 days, a full week, or more than a week?</s>

User: Just a week. 

Agent:  [INQUIRY THOUGHT] The user has clarified the duration as one week, which is a good start. Next, I should inquire about the budget to tailor the trip planning accordingly.
[INQUIRY] That's a nice amount of time to explore Paris! Now, could you tell me about your budget for the trip? Are you thinking of something economy, mid-range, or looking for a luxury experience?</s> 

User: Mid-range, nothing crazy. 

Agent:  [INQUIRY THOUGHT] The user has indicated a one-week, mid-range budget trip. Next, I should inquire about accommodation preferences.
[INQUIRY] Great choice! And where would you prefer to stay? Are you leaning towards a hotel, a hostel, or maybe an apartment rental?</s> 

User: Apartment rental. More space. 

Agent:  [INQUIRY THOUGHT] The user prefers a mid-range budget and an apartment rental for accommodation. Next, I should inquire about their activities of interest to further refine the trip planning.
[INQUIRY] Sounds cozy! What about the activities you're interested in? Are you looking forward to doing some sightseeing, immersing yourself in cultural experiences, or perhaps you're excited about shopping?</s> 

User: Sightseeing mostly. Some good food too. 

Agent:  [INQUIRY THOUGHT] User is interested in sightseeing and good food, which covers activities and dining preferences. I need to inquire about transportation needs.
[INQUIRY] For getting around Paris, would you prefer using public transport, renting a car, or having a private driver at your disposal?</s> 

User: Public transport. It's cheaper. 

Agent:  [SUMMARY THOUGHT] Based on the interaction history, I have gathered enough information to understand the user's intention for planning their trip, including duration, budget, accommodation preference, activities of interest, and transportation method. Here are the user preferences and constraints:
- Trip duration of one week
- Mid-range budget
- Preference for an apartment rental
- Interest in sightseeing and good food
- Preference for using public transport
[SUMMARY] The user intends to plan a one-week, mid-range budget trip to Paris next month, preferring to stay in an apartment rental, with a focus on sightseeing and trying good food, while using public transport for getting around.</s>
\end{lstlisting}

\subsection{Hyper-Parameter Details}
\label{apdx:method_hyperpara}

\begin{table*}[!t]
\footnotesize
\centering
\tabcolsep=0.0168\linewidth
\begin{tabular}{cccccccc}
\toprule
\textbf{Model} & \textbf{\makecell{Max \\ Length}} & \textbf{Epochs} & \textbf{\makecell{Batch \\ Size}} & \textbf{\makecell{LR}} & \textbf{\makecell{Time (h)}} & \textbf{\makecell{LR \\ Scheduler}} & \textbf{Optimizer}   \\
\midrule
Mistral-Interact & 2048       & 3         & 16                    & 1e-06         & 4.5         & Cosine                  & AdamOffload \\
\bottomrule
\end{tabular}
\caption{The hyper-parameters applied during the training of Mistral-Interact. \emph{LR} denotes the learning rate.}
\label{tab:hyperparameters}
\end{table*}

We present some key hyper-parameters in \Cref{tab:hyperparameters}. Additionally, we dynamically adjust the loss scale based on the changing training loss to prevent underflow.
Under this setting, we adapt Mistral-7B into Mistral-Interact, a powerful variant robust and effective in user implicit intention understanding during the interaction. We apply Mistral-Interact for all the experiments and evaluations.

\section{Instruction Understanding Evaluation Details}

\subsection{User Participation Details}
\label{apdx:interact_exp_user_participation}
For 108 test split tasks in IN3, we recruit eight diverse users to participate in the model-user conversation, each person in charge of several tasks. Users are all undergraduate level with different expertise. All users are asked to imagine they are doing the task given, and they could arbitrarily respond to the tested model and reflect their intentions whenever being inquired. This promotes conversation records with different tones, lengths, and styles, thus making our results diverse and representative.

The user is also asked to annotate some details along with the conversation. This includes for each round of conversation, how many queries the model inquires, how many options are provided, and whether these options are reasonable. At the end of each conversation about a task, the user also annotates how many details they offer in total, and how many are explicitly summarized finally.

We randomly sample around 5\% of the annotations to ensure the annotations are performed fairly and objectively. All the sampled annotations passed the validity check. These annotated details ease the data analysis process and calculation of some related metrics.

\subsection{Metric Measurement Details}
\label{apdx:interact_exp_metric_measurement}
We employ direct calculation, user annotations, and GPT-4 to help with our measurements. The user annotation details along with the conversation are presented in \Cref{apdx:interact_exp_user_participation}. We involve GPT-4 in the matching of missing details that the model inquires about and the missing details that IN3 presents as ground truths. Specifically, we ask GPT-4 to judge whether each piece of model inquiry could be paired with one of the missing details provided in IN3. We similarly sample 5\% of GPT-4's matching results, which pass the validity check, to ensure result accuracy. This also eases the data analysis process and promotes automation.

\subsection{Metric Calculation Details}
\label{apdx:interact_exp_metric_calculation}
\paragraph{Formalization.} We define $T$ as the set of testing tasks. Each task's conversation record involves the model's vagueness judgment $j$, multiple rounds of conversation $R$, and a final summary $s$. The annotated ground truth for this task involves the vagueness judgment $j_{truth}$, three sets of missing details $D_{truth}^{i}$ respectively of importance level $i$, {\small$1 \leq i \leq 3$}. We additionally define $T_v \in T$ as the set of tasks that the model regards as vague.

For \textbf{one round} of conversation in $R$, the assistant may inquire about multiple missing details $D = $ {\small$\{d_1, d_2, \ldots, d_n\}$}, among which only a subset $D^{i} \in D$ align with the ground truth for importance level $i$ ($D^{i} \in D_{truth}^{i}$). All the inquiries may also be accompanied by corresponding options, forming a set $P = $ {\small$\{P_{d_1}, P_{d_2}, \ldots, P_{d_n}\}$}, among which only a subset $P_r \in P$ is regarded as reasonable. Meanwhile, the user may provide response or additional information $U = $ {\small$\{u_1, u_2, \ldots, u_m\}$}, among which only a subset $U_s$ is explicitly summarized in the model's summary $s$.

\paragraph{Calculation Formulas.} We present the calculation details and formulas of each metric we provide. 
\begin{itemize}[topsep=0pt, partopsep=1pt, leftmargin=12pt, itemsep=0pt]
\item \textbf{Vagueness Judgement Accuracy}: The model's vagueness judgment $j$ is automatically determined by whether it directly provides a summary (representing clear), or makes inquiries (representing vague). The accuracy is defined as:
\begin{equation}
J_{acc} \ = \ \frac{1}{|T|} \sum_{T} (j == j_{truth})
\end{equation}
\item \textbf{Missing Details Recover Rate}: With the help of GPT-4, we disentangle and extract all the model inquired missing details $D$, and match them one by one to the ground truth missing details $D^{i}_{truth}$ of different importance level $i$. The recover rate for importance level $i$ is defined as:
\begin{equation}
RR^{i} \ = \ \frac{1}{|T_v|} \sum_{T_v} \frac{\sum_{R} |D^{i}|}{\sum_{R} |D_{truth}^{i}|} 
\end{equation}
\item \textbf{Summary Intention Coverage Rate}: The details about what is explicitly summarized ($U_s$) are annotated by the user at the end of the conversation. The coverage rate is defined as:
\begin{equation}
CR = \ \frac{1}{|T_v|} \sum_{T_v} \frac{\sum_{R} |U_s|}{\sum_{R} |U|} 
\end{equation}
\item \textbf{Options Presenting Rate}: The number of options provided for each inquired missing detail is annotated by the user during the conversation. The options presenting rate is defined as:
\begin{equation}
OP_{p} = \ \frac{1}{|T_v|} \sum_{T_v} \sum_{R} \frac{\sum_{R} |\{P_i \in P \ | \ P_i \neq \emptyset \}|}{\sum_{R} |D|}
\end{equation}
\item \textbf{Options Reasonable Rate}: The number of reasonable options is annotated by the user during the conversation. The options reasonable rate is defined as:
\begin{equation}
OP_{r} = \ \frac{1}{|T_v|} \sum_{T_v} \frac{\sum_{R} |P_r|}{\sum_{R} \sum_{P} |P_i|}
\end{equation}
\item \textbf{Average Provided Options}: The average provided options is defined as:
\begin{equation}
OP_{avg} = \ \frac{1}{|T_v|} \sum_{T_v} \frac{\sum_{R} \sum_{P} |P_i|}{\sum_{R} |D|} 
\end{equation}
\item \textbf{Average Inquired Missing Details}: The number of missing details inquired by the model in each round of conversation is annotated by the user. The average inquired missing details is defined as:
\begin{equation}
D_{avg} = \ \frac{1}{|T_v|} \sum_{T_v} \sum_{R} |D|
\end{equation}
\item \textbf{Average Conversation Rounds}: All the parameters in this metric are automatically parsed. The average rounds of conversation is defined as:
\begin{equation}
R_{avg} = \ \frac{1}{|T|} \sum_{T} |R|
\end{equation}
\item \textbf{Average Inquired Missing Details Per Round}: The average inquired missing details per round is defined as:
\begin{equation}
DPR_{avg} = \ \frac{1}{|T_v|} \sum_{T_v} \frac{\sum_{R} |D|}{|R|} 
\end{equation}
\end{itemize}

\section{Instruction Execution Evaluation Details}
\subsection{Test Data Details}
\label{apdx:agent_exp_data}
In the following, we present the ten randomly selected vague tasks within the model's ability, and their corresponding clear user goals the model summarized after interaction.\\
\makebox[\linewidth]{\rule{\linewidth}{0.4pt}}
\textit{Test Data List}
\begin{lstlisting}[basicstyle=\small\ttfamily, breaklines=true, breakindent=0em]
1. Original Vague Task: Find the latest research on diabetes treatment.
Explicit Summary Thoughts:
The user has provided specific information over the course of five interactions, which allows for a clear understanding of their request. Here are the user preferences and constraints:
- Type 1 diabetes
- Focus on lifestyle changes
- Preference for meta-analyses
- Studies from within the last two years
- Articles must be peer-reviewed and in English
Detailed User Goal: The user is seeking recent (within the last two years), peer-reviewed meta-analyses research articles in English that focus specifically on lifestyle changes related to treating Type 1 diabetes.

2. Original Vague Task: Find a recipe for homemade pizza.
Explicit Summary Thoughts:
The user has provided specific preferences over the course of four interactions, which gives a clear understanding of their intention. Here are the user preferences and constraints:
- Prefers a Margherita pizza.
- Wants a classic Italian style.
- Desires a nice thick crust.
- Requires gluten-free ingredients.
Detailed User Goal: The user is looking for a recipe for a homemade gluten-free classic Italian-style Margherita pizza with a nice thick crust.

3. Original Vague Task: Find the latest news about global warming.
Explicit Summary Thoughts:
The user's intention is now clear after four rounds of interaction, with specific preferences and constraints provided in each round. Here are the user preferences and constraints:
- Interested in policy changes related to global warming.
- Looking for updates on a regional level.
- Wants information from the past year.
Detailed User Goal: The task is to find the latest news about global warming focusing specifically on policy changes over the past year at a regional scale.

4. Original Vague Task: Search for an explanation of the String Theory.
Explicit Summary Thoughts:
The user has provided specific preferences over three rounds of interaction, which gives a clear understanding of their intention. Here are the user preferences and constraints:
- In-depth analysis of String Theory.
- Just a textual explanation without visuals or interactive content.
- Resources should be educational websites.
Detailed User Goal: The user is seeking an in-depth textual explanation of String Theory from educational websites.

5. Original Vague Task: Develop a Python script that logs daily food intake and calculates nutritional information.
Explicit Summary Thoughts:
The user has provided specific requirements over the course of five interactions, which gives a clear understanding of their needs for the Python script. Here are the user preferences and constraints:
- Track Protein intake specifically.
- Accept command-line arguments as the method of interaction with the script.
- Track protein intake over time and generate reports.
- Cater to Gluten-free dietary restrictions.
Detailed User Goal: Develop a Python script that logs daily gluten-free food intake using command-line arguments, tracks protein consumption over time, and calculates and displays protein content per meal or total daily protein intake in a file report.

6. Original Vague Task: Find a way to convert audio files from one format to another using Python.
Explicit Summary Thoughts:
The user has provided specific constraints over the course of three interactions, which gives a clear understanding of their needs for converting audio files using Python. Here are the user preferences and constraints:
- Convert MP3 to WAV format.
- Prioritize saving space in the converted file.
- Use ffmpeg as the preferred library or tool within Python.
Detailed User Goal: The task is to use Python with ffmpeg to convert MP3 audio files into WAV format while prioritizing smaller file sizes.

7. Original Vague Task: Write a summary of a given text file.
Explicit Summary Thoughts:
The user has provided specific information about the text, purpose, length, and focus through our interaction. Here are the user preferences and constraints:
- The text to be summarized is a research paper.
- The summary is for a business presentation.
- The desired length of the summary is one concise paragraph.
- Specifically highlight key arguments from the research paper.
Detailed User Goal: The task is to write a concise one-paragraph summary of a research paper's key arguments for use in a business presentation.

8. Original Vague Task: Write a Python script that sorts images based on their content.
Explicit Summary Thoughts:
The user has provided specific details over the course of four interactions that clarify their initial task, which allows for a clear understanding of their intention. Here are the user preferences and constraints:
- Recognize different scenes in images.
- Store sorted images in a database for quick search and retrieval.
- Process images using the OpenCV library.
Detailed User Goal: Create a Python script to sort images based on scene recognition, utilizing OpenCV, and storing them in a database for fast searching.

9. Original Vague Task: Write a bash command to check my system specification.
Explicit Summary Thoughts:
The user has clarified their request in two rounds of interaction, specifying the type of system information they want and the level of detail required. Here are the user preferences and constraints:
- Check memory usage.
- Provide a comprehensive list with details.
Detailed User Goal: The task is to write a bash command that provides a comprehensive list with detailed information on the user's system memory usage.

10. Original Vague Task: Set up a study plan for my upcoming law school exams in two months.
Explicit Summary Thoughts:
The user has provided specific details over the course of four interactions that allow for a clear understanding of their study plan needs. Here are the user preferences and constraints:
- Focusing on Criminal Law, which is a struggle area.
- Exams will be mainly essay-based.
- Prefers to study alone.
- Has approximately 5 hours per day available for studying, with two months until exams.
Detailed User Goal: Create an intensive Criminal Law study plan focusing on essays for a law school examinee who prefers independent learning, allocating 5 hours daily over two months.
\end{lstlisting}

\subsection{Metric Measurement Details}
\label{apdx:agent_exp_metric}
We evaluate all the agent's performance based on the specific and detailed task goal after the interaction. For all three metrics, we asked the same user who had a conversation with Mistral-Interact about this task to do an additional round of annotation, respectively about each subtask and milestone's generality and necessity. The necessity is judged upon whether the description entails any given details in the user's intention, while the generality is judged upon whether the corresponding execution meets the user's detailed and tailored intentions.